# From Whole-slide Image to Biomarker Prediction: A Protocol for End-to-End Deep Learning in Computational Pathology

Omar S.M. El Nahhas (1), Marko van Treeck (1), Georg Wölflein (2), Michaela Unger (1), Marta Ligero (1), Tim Lenz (1), Sophia J. Wagner (3, 4), Katherine J. Hewitt (1), Firas Khader (5), Sebastian Foersch (6), Daniel Truhn (5), Jakob Nikolas Kather (1, 7, 8, 9)

(1) Else Kroener Fresenius Center for Digital Health, Medical Faculty Carl Gustav Carus, Technical University Dresden, Dresden, Germany.
(2) School of Computer Science, University of St Andrews, St Andrews, United Kingdom
(3) Helmholtz Munich – German Research Center for Environment and Health, Munich, Germany
(4) School of Computation, Information and Technology, Technical University of Munich, Munich, Germany
(5) Department of Diagnostic and Interventional Radiology, University Hospital Aachen, Aachen Germany
(6) Institute of Pathology - University Medical Center Mainz, Mainz, Germany
(7) Pathology & Data Analytics, Leeds Institute of Medical Research at St James's, University of Leeds, Leeds, United Kingdom
(8) Medical Oncology, National Center for Tumor Diseases (NCT), University Hospital Heidelberg, Heidelberg, Germany
(9) Department of Medicine 1, University Hospital and Faculty of Medicine Carl Gustav Carus, Technische Universität Dresden, Dresden, Germany

# Abstract

Hematoxylin- and eosin (H&E) stained whole-slide images (WSIs) are the foundation of diagnosis of cancer. In recent years, development of deep learning-based methods in computational pathology enabled the prediction of biomarkers directly from WSIs. However, accurately linking tissue phenotype to biomarkers at scale remains a crucial challenge for democratizing complex biomarkers in precision oncology. This protocol describes a practical workflow for solid tumor associative modeling in pathology (STAMP), enabling prediction of biomarkers directly from WSIs using deep learning. The STAMP workflow is biomarker agnostic and allows for genetic- and clinicopathologic tabular data to be included as an additional input, together with histopathology images. The protocol consists of five main stages which have been successfully applied to various research problems: formal problem definition, data preprocessing, modeling, evaluation and clinical translation. The STAMP workflow differentiates itself through its focus on serving as a collaborative framework that can be used by clinicians and engineers alike for setting up research projects in the field of computational pathology. As an example task, we applied STAMP to the prediction of microsatellite instability (MSI) status in colorectal cancer, showing accurate performance for the identification of MSI-high tumors. Moreover, we provide an open-source codebase which has been deployed at several hospitals across the globe to set up computational pathology workflows. The STAMP workflow requires one workday of hands-on computational execution and basic command line knowledge.

# Introduction

Routine histopathology slides of solid tumors are available at scale and harbor clinically actionable information[1]. Analyzing digitized whole slide images (WSIs) is a complex task that requires multidisciplinary collaboration between clinicians and engineers. Unfortunately, bringing these experts together is challenging, thus hindering progress in the field. Deep learning has been used to process these WSIs[2], to diagnose[3,4] and subtype tumors[5,6], but also to extract more abstract properties, such as information about the tumor prognosis[7,8] and treatment response[9,10]. Deep learning-based biomarkers that extract quantitative prognostic or predictive information from routine pathology slides have the potential to be transformative for precision oncology. This could lead to improved treatment for patients, rather than just optimizing workflows[11]. In the last few years, dozens of academic studies have shown that such biomarkers can be extracted from routine tissue slides in multiple tumor types[12], such as lung cancer[13,14], colorectal cancer[15–18], liver cancer[19], breast cancer[20,21], and renal cell cancer[22]. These studies used a weakly-supervised prediction approach, meaning they used deep learning to predict a label which is defined for a slide or a patient without the need for annotations at pixel-level.

Computational pathology research projects share a common computational approach[23]: they begin with a digital histopathology slide, divide it into thousands of image tiles, extract features from each tile, and employ a machine learning model to combine these features at the slide level for a specific prediction objective. The repeated redevelopment of core computational pathology workflows[13,24–27] not only obstructs advancements in the field through time lost on standard workflows, but also causes inefficiency and risks low reproducibility due to the complex nature of the computational pipeline[28]. Furthermore, as computational techniques evolve, they tend to become more intricate and intertwined within complex source code, rendering them less accessible to clinicians and raising the threshold for their adoption in research.

To address this issue, we put forth an end-to-end protocol for solid tumor associative modeling in pathology (STAMP), that streamlines the analysis of WSIs and enables medical and technical experts to work together effectively. Our protocol provides a comprehensive framework for going from WSI to patient-level biomarker prediction through a step-by-step guide accompanied by layperson explanations of the algorithmic- and implementation details, facilitating communication and collaboration between experts with different backgrounds. We have developed an open-source software pipeline that is flexible, modular and allows for a low barrier approach to weakly-supervised deep learning in pathology. The modularity facilitates independent and asynchronous execution of each step within the STAMP workflow, mirroring the operational structure of computational pathology projects conducted with real-world data. This pipeline was used in several clinical studies led by both clinicians and engineers, including gastric cancer[29–31], colorectal cancer[32,33], breast cancer[34], brain cancer[35], and pan-cancer[36–38] studies. It has shown state-of-the-art performance in weakly-supervised computational pathology for clinically relevant prediction tasks. By following our protocol, medical and technical experts can speak the same language and utilize the same tools to answer highly relevant clinical questions, enabling computational pathology projects to be initiated at scale and potentially leading to new discoveries towards improved patient outcomes.

## Overview of the protocol

The protocol consists of five overarching steps: 1) problem definition, 2) data preprocessing, 3) deep learning, 4) model evaluation, and 5) clinical translation (**Fig. 1**). The novelty of the protocol lies within its ease of use for clinical researchers to set up a complete computational pathology project, as well as utilizing the latest deep learning techniques within a general applicable workflow to map the non-linear phenotypic-genotypic relationships directly from routine WSIs. Specifically, using a state-of-the-art pretrained self-supervised learned histology feature extractor[39] and transformer architecture[33] (**Fig. 2**). To exemplify the protocol, the phenotypic features of colorectal cancer (CRC) WSIs are mapped to the genotypic features of mismatch repair (MMR) genes related to microsatellite instability (MSI). This cancer type and clinical endpoint were chosen due to their known strong phenotypic-genotypic correlation[15,32,33,40]. This protocol, however, has shown to be broadly applicable to numerous solid tumor types and various categorical or continuous genetic biomarkers[29–31,36–38]. The protocol is accompanied by an open-source codebase (available at https://github.com/KatherLab/STAMP). No code customization is required to use this protocol on new datasets and tasks, only a change of values inside the accompanying configuration file.

It is important to emphasize that our protocol exclusively enables associative modeling between input features and outcome targets, rather than facilitating causal inference. In other words, it supports the identification of relationships between tissue phenotype and biomarkers without determining whether specific features directly cause particular outcomes. Causality in digital medicine is complex and an ongoing area of research, demanding a multidisciplinary environment encompassing domain expertise for accurate assumptions[41].

## Expertise needed to implement the protocol

The protocol is composed of three distinct phases: orchestration, execution, and interpretation (**Fig. 1)**. Orchestration involves problem formulation and data curation (**Steps** 1-6). It requires clinical expertise to ask the right medical questions, eliminate low-quality WSIs, and transform raw endpoint measurements into a patient-level label. The execution phase is primarily focused on technical aspects, ranging from image preprocessing to modeling and technical evaluation (**Steps** 7-17), which should be supervised by an engineer with a solid understanding of image processing and deep learning techniques. The interpretation phase, which encompasses clinical evaluation and explainability (**Steps** 18-25), is a collaborative effort between clinicians and engineers. During this phase, clinicians concentrate on the medical relevance and interpretability of the results, while engineers focus on technical aspects, such as overfitting, which may affect the results. By leveraging the multidisciplinary expertise of clinicians and engineers across all three phases, the protocol can be effectively executed and provide meaningful results. To ensure effective interdisciplinary collaboration for this protocol, a general guideline is to have a clinician lead the orchestration and interpretation phase, while an engineer leads the execution phase. Additionally, a glossary of common terms is provided to facilitate communication between clinicians and engineers (**Box 1**).

## Dataset requirements and limitations

Implementation of our protocol requires WSIs, accompanied by related clinical information and molecular data. Here, we used the open-source datasets from The Cancer Genome Atlas[42] (TCGA) and Cancer Proteomics Transcriptomic Tumor Analysis Consortium[43] (CPTAC) efforts. The minimum number of patients required for AI analysis of WSIs is dependent on various factors (**Fig. 3**). For example, the size of the dataset depends on the strength of the correlation between tissue phenotype and biomarkers, and larger datasets may be necessary for problems with weak correlations between phenotype and genotype. On the other hand, smaller datasets may be sufficient for problems with strong correlations. Additionally, the number of positive cases in the dataset is a crucial factor to consider. For problems that involve rare mutations or events, larger datasets are necessary to ensure adequate representation of these positive cases. Conversely, for more common problems, smaller datasets may suffice. Finally, the quality and size of the tissue plays an important role. These factors must be carefully considered when determining the appropriate dataset size for AI analysis of WSI in computational pathology.

Limitations and assumptions of the protocol presented in this study include the requirement of at least one WSI per patient, as well as a label for each patient. It should be noted that the resulting models' performance may be limited in cases of low data availability (<100 patients), in instances of severe class imbalance between positive and negative cases (<5 positive cases), or in instances with highly skewed distribution for continuous scores. Our codebase that accompanies the protocol has a limitation in terms of the supported file types for WSIs. Specifically, it only supports file types that are compatible with the OpenSlide library[44]. Not all image file types are supported, and this may affect the applicability of our protocol to some datasets. The supported file types include SVS, MRXS, NDPI, and other file types which can be transformed into pyramidal TIFF files. It is important to consider file type compatibility when selecting datasets to apply our protocol. Moreover, the protocol assumes access to sufficient computational resources to preprocess the WSIs, train models, and compute heatmaps. The workflow can only be initiated from a command line interface (CLI), therefore requiring at least a basic understanding of running a command in such an interface. Altering the models' hyperparameters requires knowledge of Python programming, as not every parameter can be adjusted through the CLI.

It should be noted that this protocol is meant as a lean and low-barrier entry for weakly-supervised deep learning in computational pathology. Many extensions have been added over time, where the user is able to choose between pretrained models for feature extraction, use different models for downstream finetuning tasks, opt between classification and regression, perform multi-task learning, and choose different ways of generating spatial attention heatmaps. For the purpose of simplicity, those features are not present in the main protocol, but are made available open-source (https://github.com/KatherLab) for advanced users. Moreover, the software utilized for problem formulation and clinical translation are not included in the STAMP computational workflow, as this is a process which is often performed using tools such as Excel, programming languages such as R, or pathology slide viewers such as QuPath[45].

## Comparison with other protocols

In recent years, several general-purpose enabling software supporting WSI deep learning protocols have been developed, such as PyTorch[46], MONAI[47], OpenSlide[44], LibVips[48] and RAPIDS cuCIM (https://rapids.ai). These tools form the basis for loading WSIs and providing an environment to create deep learning models. Building on top of aforementioned tools are the domain-specific adaptable software, such as HistoQC[49], FastPathology[50], TIAToolbox[51] and CLAM[52]. HistoQC is a software tool used in digital pathology to assess the quality of WSIs slides in an automated manner, but does not provide tools for further computational analysis related to biomarker prediction. The FastPathology platform provides a platform with a visual interface capable of supervised deep learning for segmentation and object detection, but does not provide modular software for asynchronous execution of processing steps. The TIAToolbox software provides a toolbox for processing and modeling of WSIs, offering a large range of tools including WSI pre- and post-processing, instance segmentation and weakly-supervised deep learning. Similarly, the CLAM software encompasses several tools for pre- and post-processing of WSIs, but is focussed on weakly-supervised deep learning. Furthermore, there has been a noticeable increase in the development of proprietary single-application software within the medical field[53]. This involves transforming open-source domain-specific adaptable software applications into products that have received certification from regulatory authorities like the US Food and Drug Administration (FDA) and the European Conformité Européenne for In Vitro Diagnostic (CE-IVD) standards, such as a tool for prediction of MSI status in CRC[54] which is developed with a similar philosophy as proposed in prior research[15] and is analogous to the STAMP computational workflow.

A recent review identified three general weaknesses of currently available computational pathology workflows[28]: the lack of experiment reproducibility using the code, the lack of real-world translation of results, and the lack of supporting documentation to reuse the code. The review underlines that the currently available workflows are missing supporting documentation alongside the codebase, hindering researchers without a computational background to reuse the provided software. Aforementioned workflows have in common that they are technically robust techniques which require programming knowledge to use, but lack thorough documentation for non-computational researchers to interact with their software and conceptualize a computational pathology study.

Our protocol provides lean software for WSI processing focused on weakly-supervised deep learning in an end-to-end and generally-applicable manner, which requires no programming knowledge to utilize. We contribute to existing software by putting emphasis on lowering the entry barrier for use by clinicians (**Fig. 4)**, while facilitating the use of state-of-the-art computational methods which have been utilized in a plethora of clinician-led[29,30,32,35] and engineer-led[31,33,34,36] studies . Moreover, our protocol goes beyond providing a software workflow by including guidance on problem formulation and clinical translation, providing a guideline for end-to-end computational pathology projects using real-world data. Through our protocol, we aim to enable fruitful collaborations between clinicians and engineers to work on research questions in computational pathology, promoting the development of more certified diagnostics products that lower the burden on the healthcare system and improve patient outcomes.

# Materials

## Data

Histopathology slides and genomics data from TCGA and CPTAC were used to train and validate the models. The slides for TCGA are available at https://portal.gdc.cancer.gov/. The slides for CPTAC are available at https://proteomics.cancer.gov/data-portal. The molecular and clinical data for TCGA and CPTAC used in the experiments are available at https://github.com/KatherLab/cancer-metadata.

## Hardware

Preprocessing is the heaviest operation in the protocol and requires a large amount of random-access memory (RAM), depending on the size of the WSI. The entire process can be run on central processing units (CPU), requiring a minimum of approximately 50 gigabytes (GB) of system RAM, being strongly dependent on the WSI size and chosen microns-per-pixel (MPP) ratio. No graphics processing units (GPUs) are required to run STAMP, but speed improvements can be made by using GPUs. If a GPU is used in preprocessing, at least 10 GB of video RAM (VRAM) is necessary, whereas a GPU used in modeling requires upwards of 40 GB of VRAM due to the computationally demanding transformer architecture.

## Software

The Python software packages for the protocol can be installed all at once using the STAMP installer (**Box 2**), as also described in the *`README.md`* file (https://github.com/KatherLab/STAMP). The required software is listed in the *`pyproject.toml`* file and is intended for usage in Python 3.10 or higher (**Suppl. Table 1**). Note that these packages are the key building blocks and are dependent on other supportive packages, which will trigger a series of installations to satisfy the dependencies. Therefore, this list is not an exhaustive overview of every package, but can be obtained by using CLI commands *`conda list`* or *`pip list`* inside your environment. The configuration for the entire protocol is set by the user in the configuration file, *`config.yaml`* (https://github.com/KatherLab/STAMP/blob/main/config.yaml). The STAMP workflow has been developed for Linux operating systems, but can be used on any operating system using the provided containerized environment on GitHub.

# Procedure

The software was developed to involve as little additional programming or scripting for the user as possible. Consequently, the user can fill in all paths, variables and settings in the configuration file, which will serve as the configuration for each step described below that requires interaction with the STAMP software (**Box 3**). The configuration file should be saved after every change applied to it before running the `stamp` CLI commands.

## Orchestration: Formal problem definition

1) ***Define the clinical use-case***: Define the clinical hypothesis to be tested and the target to be predicted. The pipeline for this protocol is intended for weakly-supervised problems only. Patient-level biomarker labels are used for training the model, without any annotations of the WSI. The main use-case of this protocol is the detection of biomarkers directly from WSIs.

2) ***Assess the quantity of data***: Identify how many patients with corresponding WSIs and biomarker labels are available. A patient can have multiple H&E-stained slides.

3) ***Define the inclusion and exclusion criteria***: Visually evaluate the quality of the WSIs using a pathology slide viewer of choice, such as QuPath[45]. It is important to identify slides that have tumor tissue, contain pen markings, are biopsies or resections, and are either a primary or metastasized tumor. The metadata of the WSIs should contain the microns per pixel (MPP) value, which is used for the digital loading and processing of the slides. Define which patients and corresponding slides are used for subsequent analysis based on the data quality. Specifically, the quality measures regarding availability of the target label, presence of dirt and pen markings, the presence of tumor tissue, the type of surgical sample (biopsy or resection), the type of tissue (primary tumor or metastasis) and the availability of the MPP in the WSIs' metadata[23].

4) ***Determine desired slide resolution***: Define which slide resolution is required for the use-case defined in **step** 1. The slides are usually scanned at 40x or 20x magnification, approximately corresponding to an MPP value of 0.25 and 0.5, respectively. The magnification to MPP ratio is scanner dependent. The pipeline allows for downscaling to lower resolutions, but does not allow for upscaling to resolutions beyond the highest available representation in the pyramidal slide files' metadata. With the default values, the analyzed tile resolutions are equivalent to approximately 10x magnification (1.14 MPP), empirically optimized for modeling performance and required computational resources.

5) ***Transform the data***: First, decide whether the biomarkers to predict require transformations before modeling based on **Steps** 1-4. Transformations could include the removal or merging of categories within a biomarker, binning continuous targets, or merging several biomarkers. Second, decide whether the WSIs should be stain normalized[55]. Stain normalization is commonly used for H&E-stained slides originating from different hospitals, applied to reduce batch effects[56]. The transformations of the biomarker data make the subsequent analysis more manageable, as this is a way of introducing domain-knowledge which discards abundant or contradicting information before it is seen by an algorithm.

6) ***Define the validation strategy***: First, establish the patient allocation strategy to validate the hypothesis defined in **step** 1. The validation strategy requires the inclusion and exclusion criteria defined in **step** 3 to identify the number of patients available for the internal training cohort and external testing cohort. Following best scientific practice in the field of computational pathology and machine learning, this protocol assumes the availability of an external cohort that can be used for testing the model's generalizability. Second, define the statistical tests and metrics which enable testing of the hypothesis. Establish clear criteria and define value ranges for metrics, as well as specify the corresponding statistical tests, to either accept or reject the hypothesis. This includes expected presence of certain cells or genomic alterations predominantly belonging to the categories of the biomarker.

## Execution: Data preprocessing

7) ***Set up computing resources:*** First, follow the installation instructions, and finalize the installation of the STAMP workflow (**Box 2**). Second, ensure that a configuration file exists, such as the aforementioned "*config.yaml*" example file. Next, make the appropriate computing resources available for preprocessing the WSI data as indicated in the hardware section. Preprocessing of a WSI for DL purposes is a data- and computational intensive procedure. The computational demand is dictated by the size of the gigapixel WSI, as well as the chosen parameters for the MPP. Reducing the MPP leads to a higher resolution, resulting in a larger amount of tiles, which heavily increases the computational resources needed.

8) ***Set up preprocessing configuration***: Open the configuration file, and insert the required arguments for the *preprocessing* section (**Table 1**). The predefined settings for preprocessing are recommended to be kept as the default values to reproduce the experimental results.

9) ***Extract features from WSI***: Run the preprocessing pipeline using the configuration from **step** 8 with the CLI command `stamp preprocessing`. This triggers a chain of processes which loads the WSIs into memory at a magnification defined by the MPP, tessellates the slide into *n* tiles of 224x224 pixels, removes tiles which contain too little tissue or are blurry through Canny edge detection[57], optionally normalizes the H&E staining color distribution of the *n* tiles according to Macenko's method[55], runs inference with the feature extractor model on each tile to obtain its feature vector, and then concatenates the feature vectors of all *n* tiles of one slide into one large feature matrix. By default, the preprocessing utilizes a robust[58] model which has been trained on 14 million patches from 32,000 publicly available slides across various cancer types, CTransPath[39]. The feature matrix for each slide has a dimensionality of *n* x 768 for CTransPath, where the *n* tiles vary per slide. The preprocessing step has two key outputs, the intermediate products of the WSIs and a feature vector file. The intermediate products are images which can be visually inspected to determine the correct behavior of the preprocessing procedure. The extracted features are the final product, which is used later in the modeling section to build a classification DL model.

10) ***Assess image preprocessing***: Analyze the intermediate products of the preprocessing pipeline which are stored in the directory given as input for the *cache_dir* argument from the *preprocessing* section in the configuration file. First, observe the CLI and identify whether the script has successfully run. A successful run finishes with a summary regarding the number of slides that were processed, and the total runtime of the preprocessing. Second, scroll through a sample of the folders named after the WSI that was analyzed, and perform a quality control of the files within each folder: `slide.jpg`, `canny_slide.jpg` and `norm_slide.jpg`. These image files contain the WSI loaded at the desired MPP, the partially preprocessed image after rejecting blurry and background tiles, and the fully preprocessed image with stain normalized tiles, respectively. Each non-rejected tile present in the `norm_slide.jpg` file will result in a feature vector that will be concatenated into a feature matrix for the entire slide. Check whether the quality of the processed WSIs are acceptable to test the clinical hypothesis as defined in **Steps** 1-6, especially regarding the MPP, tile size and the presence of slide artifacts like air bubbles, dirt and penmarks. This protocol allows for non-tumor-containing tiles to be kept, as the downstream classification tasks learn relevant from non-relevant tiles when provided with sufficient samples, including penmarks[32]. It is, however, recommended to use slides where non-essential information is kept to a minimum. Note that image features of a cohort only have to be extracted once for each desired resolution, and can be reused for other downstream classification tasks.

# Execution: Modeling

11) ***Define slide table:*** Create a table which links a patient identifier to the filename of the extracted slide features[23] (**Box 4**).

12) ***Define clinical table***: Create a table which links a patient identifier to the biomarker data[23] (**Box 5**).

13) ***Define data splits:*** Determine how the data should be split for modeling (**Box 6**). The default splitting mechanism is 5-fold cross-validation (**Suppl. Fig. 1**) and a single 80-20 split for the final model training.

14) ***Set up modeling configuration:*** Open the configuration file, and insert the required arguments for the *modeling* section (**Table 2**). The arguments in the modeling section are used for cross-validation, final model training, and deployment of the model on an external cohort, depending on the CLI STAMP command that is used. At this step, provide the arguments required for cross-validation.

15) ***Train cross-validated models:*** Run the modeling pipeline with cross-validation using the configuration from **step** 14 with the CLI command `stamp crossval`. By default, cross-validated training uses the 5-fold data splits as described in **step** 13. This process results in the creation of 5 distinct models, all built with the same architecture but exhibiting differences in their learned parameters due to the diverse training data. This allows the model architecture performance to be measured under various data conditions.

16) ***Evaluate cross-validation***: Open the configuration file*,* and insert the required arguments for the *statistics* section (**Table 3**). Measure the performance of the cross-validated models using the CLI command `stamp statistics`. The area under the receiver operator characteristic (AUROC) and area under the precision recall characteristic (AUPRC) are two computed metrics which are, preferably, as close to 1 as possible, indicating the quality of the classification. Simultaneously, the 95% confidence interval (CI) is calculated, where a small 95%CI indicates greater robustness in performance across the various cross-validated models. Note that additional metrics defined in **Step** 6 can be calculated manually using the patient predictions and are not included in the STAMP workflow by default.

17) ***Train the final model***: Open the configuration file*,* and insert the required arguments for the *modeling* section (**Table 2**). At this step, provide the arguments required for full-training. It is suggested to choose a different output directory for the *output* argument in the modeling section, to not interfere with the results from the cross-validation in **Steps** 15 and 16. Run the modeling pipeline to train the final model using CLI command `stamp train`*.* This process results in the creation of a single model file (*.pkl*) which has been trained on 100% of the cohort, thus requiring an external cohort to measure its performance in predicting the biomarker.

# Execution: Evaluation

18) ***Set up external validation configuration***: Open the configuration file*,* and insert the required arguments for the *modeling* section (**Table 2**). At this step, provide the arguments required for model deployment on an external cohort. Repeat **Steps** 7-10 to preprocess the external cohort and obtain the features vectors which are used for validation of the final model. Repeat **Steps** 11 and 12 to obtain the corresponding slide- and clinical table for the external cohort.

19) ***Deploy the final model***: Run the modeling pipeline to deploy the final model on an external cohort using the configuration from **step** 18 with the CLI command `stamp deploy`*.* When deploying the model onto new data, the model is not trained or updated. The output of the model deployment on the external cohort is a single file containing the predictions for each patient.

20) ***Evaluate the final model:*** Obtain the metrics and statistics for the final model's performance on the external cohort. First, update the configuration file with the single file with patient predictions (**Table 3**). Second, use the CLI command `stamp statistics`. The predictions are sampled 1,000 times to recalculate the metrics, yielding a 95%CI for the external validation cohort modeling performance.

21) ***Generate spatial interpretability map:*** Use the gradients of the final model to generate a prediction heatmap. First, inspect the patient prediction file used in **step** 20 to calculate the metrics. Here, the prediction scores for each patient are listed and ranked by loss - the most accurate predictions are at the top of the file. Second, select patients to analyze the spatial interpretability map from. It is recommended to choose patients with an extreme value of the prediction score, i.e., the scores closest to 1. A more extreme prediction score value indicates a stronger signal was captured by the model, which helps with human interpretability of the heatmaps. Second, select the slide names belonging to the patients, observed in the slide table file from **step** 18. Open the configuration file*,* and insert the required arguments for the *heatmaps* section (**Table 4**). Generate the slide heatmaps using the CLI command `stamp heatmaps`. This results in a heatmap showing the spatial relationship of the most important tiles for the model's slide-level prediction, adding interpretability to the model's decision-making. Moreover, the desired amount of influential tiles for decision-making are stored as separate image files with the score and coordinates in the naming.

22) ***Review the model's technical performance:*** First, investigate the model's performance from the metrics and statistics calculated in **step** 20. The AUROC and AUPRC with 95% confidence interval (CI) are desired to be as close to 1 within the range of [0,1], with the 95% CI not crossing the point of insignificance. The point of insignificance lies at 0.5 for the AUROC, and is dependent on the data distribution of the target label for the AUPRC, indicating the predictions are random when crossed. An acceptable performance is dependent on the problem defined in **Steps** 1-6, where predicting MSI status is known to be accurately predicted[15,32,33], whereas predicting survival is a relatively more complex problem to model[59,60], which is reflected in the performance metrics. Second, assess whether the model pays attention to parts of the WSI which contain tissue, and it does not focus on dirt, penmarks, or other artifacts present on the slide after preprocessing. If the model does focus on parts of the WSI which are not related to the tissue, the data needs to be better curated (**Steps** 2-5), or the model requires more samples to avoid overfitting.

## Interpretation: Translation

23) ***Conduct concordance analysis:*** Use the final model's predicted scores of the external cohort to measure concordance with clinicopathological and biological variables which are relevant to test the hypothesis as defined in **Steps** 1 and 6. The aim is to find patterns in the model predictions which align with readily known clinical concepts, which hint towards the model's capability of learning relevant biology for its predictions. The software for performing a concordance analysis is not included in the STAMP workflow.

24) ***Review the model's clinical utility:*** Let a pathologist review the spatial interpretability map and corresponding tiles which influenced the model's decision-making from **step** 21. Link the findings of the pathologist and the concordance analysis in **step** 23 to literature of prior studies to discover if the developed biomarker is in accordance with known biological and medical concepts. Finally, the guidelines defined by STARD-AI[61] or TRIPOD-AI[62] can be used to formally evaluate and report the model's clinical utility.

25) ***Calculate prognostic capability:*** Use the predicted labels of the external cohort by the final model to stratify risk groups. Visualize the stratification of risk groups using Kaplan Meier curves. The predicted labels are used to separate patient groups for which the survival variable is plotted. Calculate statistical significance of different survival between the stratified risk groups using the log-rank test. Use the predicted labels of the external cohort by the final model and covariates (such as tumor stage, age and sex) to perform the multivariable Cox proportional-hazards test. This yields a Hazard Ratio (HR) and corresponding p-value. Statistical significance of the HR is achieved when the 95% CI does not overlap with HR=1. Models that produce HR values further away from 1 indicate a more pronounced effect and are considered to have greater prognostic value. The software for calculating prognostic capabilities is not included in the STAMP workflow.

# Troubleshooting

The most common troubleshooting in this protocol is software related. Usage of the software in the provided container minimizes the risk of faulty installations, which eliminates software dependency related bugs to cause unwanted behavior. Nonetheless, we have identified common scenarios from users around the world where the code could not process the data as desired and summarized those (**Table 5**). Unidentified errors in the code can be submitted to our issues page on GitHub (https://github.com/KatherLab/STAMP/issues) to further increase the robustness of the computational workflow.

# Timing

The estimated duration of the protocol is heavily reliant on the defined problem, the amount of samples, and the available computational resources (**Table 6)**. For the clinician-led steps, it is assumed a project idea is conceptualized from scratch, but data has been readily collected. For the engineer-led steps, it is assumed a Linux system has been readily set-up on which computations can be run.

# Anticipated results

After preprocessing the WSIs, 599 slides and 372 slides remain for the TCGA and CPTAC cohorts, respectively. For the actual training and deployment, which depends on the available slides and biomarker information, 444 patients were used for the internal training cohort TCGA, and 105 patients for the external validation cohort CPTAC.

Performing a 5-fold cross-validation strategy for model exploration, the Transformer models reach a mean AUROC and 95%CI of 0.84 [0.75-0.93] and mean AUPRC and 95%CI of 0.58 [0.39-0.77] for the prediction of MSI-high (MSI-H) (**Fig. 5A-B)**. This indicates robust predictive capabilities of the model on the internal training cohort. However, to measure generalizability and robustness, the model should be deployed on an external cohort. Therefore, a single Transformer model is trained on the full dataset of the TCGA training cohort, which is then deployed on the external CPTAC cohort. The deployed model shows accurate performance on the external CPTAC cohort, reaching an AUROC with bootstrapped 95%CI of 0.85 [0.74-0.94] and AUPRC with bootstrapped 95%CI of 0.68 [0.55-0.87] (**Fig. 5C-D)**. This indicates the model has learned generalizable features which perform similarly in both the internal and external test cohort. Note that the training procedure is of a stochastic nature, which might result in slightly different numbers when running experiments. The provided confidence intervals serve as a guideline for the expected range of results when running multiple experiments on the same data with the same settings, and is in line with the reported results in the reference paper[33]. The concordance analysis reveals that the fraction of genome altered is significantly lower among MSI-H predicted tumors than MSS predicted tumors ($p<0.0001$), which is in line with a key prior study[63]. Moreover, the mutation count and mutation rate of the MSI-H predicted tumors are significantly higher than the MSS predicted tumors ($p<0.0001$), which is in accordance with the concept of MSI-H status being a pattern of hypermutation[64]. These data show indications of the model having learned clinically known concepts surrounding MSI status.

To further assess the model's capability of having learned clinically relevant concepts, heatmaps of the external CPTAC-CRC cohort are generated. The top tiles indicate the key influential tiles which were used to predict MSI-H (88% confidence) and to predict MSS (12% confidence) (**Fig. 6A**). A pathology resident in-training, KJH, assessed that high-confidence predictions for MSI-H tiles tend to contain mucin and have a presence of signet rings, which are tissue characteristics predominantly associated with MSI-H status[65]. Moreover, these findings are in line with a prior study on a commercially-available MSI status prediction model in computational pathology[54], suggesting our model's correct identification of the phenotype corresponding to a MSI-H tumor in CRC. Low confidence predictions, such as 12% for MSS, can contain tissue in the top tiles that is deemed of no use for diagnosis and should be handled accordingly.

To assess the prognostic value of the predicted biomarker, the CPTAC cohort is stratified by ground-truth and predicted MSI status with disease-free survival (DFS) as endpoint. The analysis is limited to CRC patients from stage I, II and III, due to only having a single datapoint available for stage IV. From the 105 patients in the CPTAC-CRC cohort, 18 observations were deleted due to missing DFS data, resulting in an analysis with n=80 patients with 7 events in total. The stratification of stage I-III patients from CPTAC-CRC using the ground-truth MSI status yields no significant stratification (p=0.80) (**Fig. 6B**), whereas the stratification using the predicted MSI status does yield a significant stratification (p=0.04) (**Fig. 6C**), as evaluated through the log-rank test. When measuring the prognostic value of MSI status together with age, sex and tumor stage covariates, the covariates (p>0.05), ground-truth (p=0.84) and predicted (p=0.08) MSI status are not yielding significant hazard ratios, as measured by the multivariable Cox proportional-hazards model. Note that the prognostic analysis serves as an example on open-source data to exhibit the utility of the predicted biomarkers, but is limited due to the low amount of survival events. The low sample size is also a reason for the counter-intuitive risk-group assignment of the predicted MSI-H status, which is found to have a better prognosis in early-stage CRC[66]. Therefore, no conclusive clinical findings should be drawn from this example use-case of the prognostic value assessment through the STAMP protocol.

These data show the successful application of the STAMP workflow for a clinically relevant research question. This protocol aims at accelerating biomarker discovery by democratizing the best practices in weakly-supervised computational pathology, enabling research towards better precision oncology at scale.

# Figures

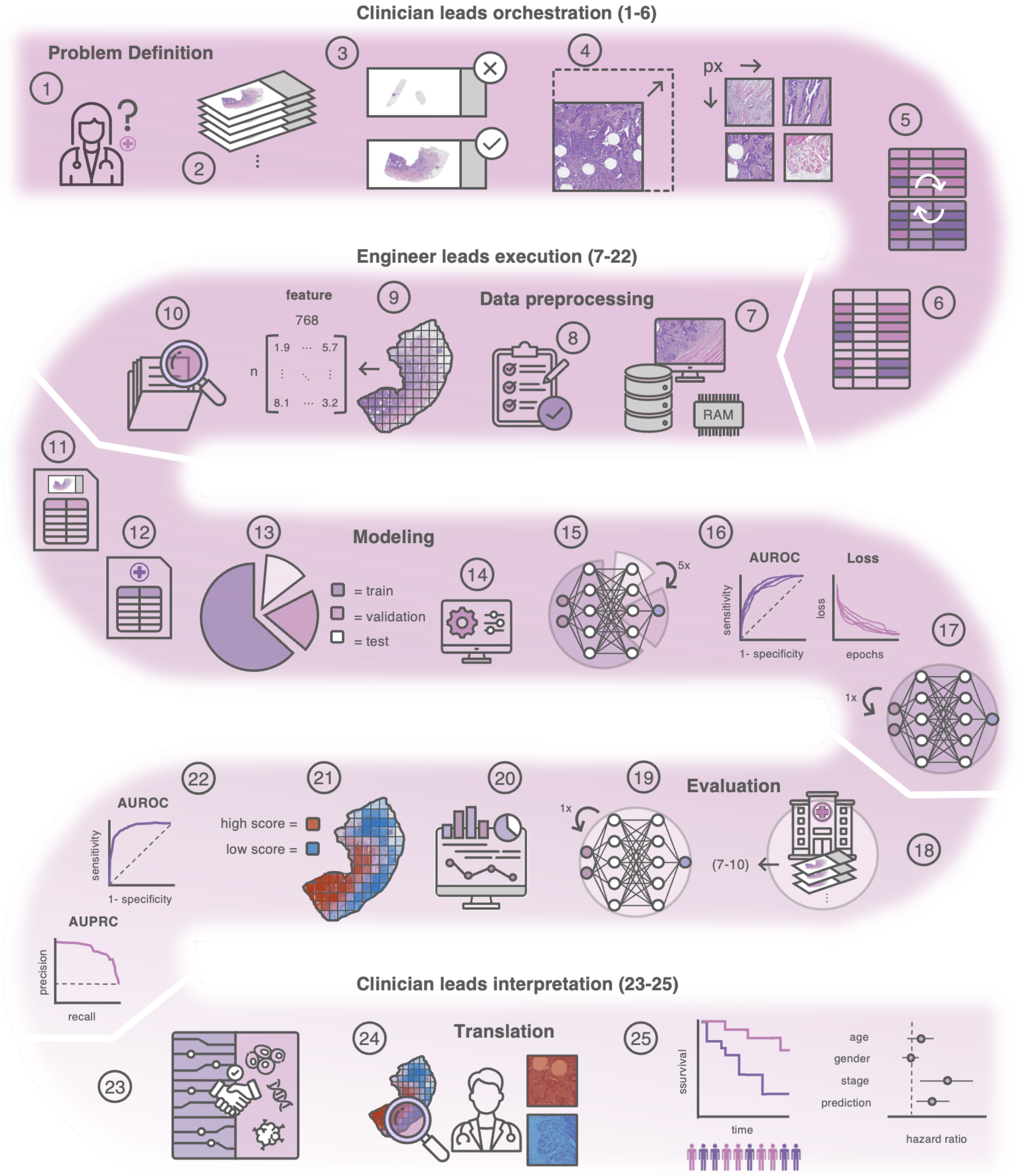

**Figure 1: Conceptual overview of the protocol.** The protocol starts with the definition of the clinical hypothesis and prediction target for the model **(1)**, leading to data collection, which includes assessing the number of slides **(2)** and the quality of the data **(3)** to establish inclusion and exclusion criteria. Subsequent steps involve defining the desired image resolution **(4)**, the cleaning and transforming of the prediction categories of the target biomarker **(5)**, and defining the strategy to validate the clinical hypothesis **(6)**. Then, evaluation of computer resources **(7)** and setting up the configuration file **(8)** is required to perform image pre-

processing including normalization, tessellation and feature extraction **(9)**. It is recommended to perform an assessment of the preprocessed images **(10)** prior to the modeling steps. Setting up the modeling configuration involves defining slide and clinical files **(11-12)**, defining the data splits for modeling **(13)**, and training the model on the selected validation approach **(14-15)**. The model performance is then evaluated on the cross-validation **(16)** and retrained using all data to obtain a final model **(17)**. Next, the model is deployed and evaluated on an external cohort **(18-20).** For explainability purposes, spatial interpretability maps are generated **(21)**, with which the model's technical performance is reviewed **(22)**. The associations between the model and related clinical data **(23)** together with review of the spatial heatmaps assist the clinician to assess the clinical relevance of the model's predictions **(24)**. Finally, the prognostic capability of the predicted biomarker is calculated **(25)**.

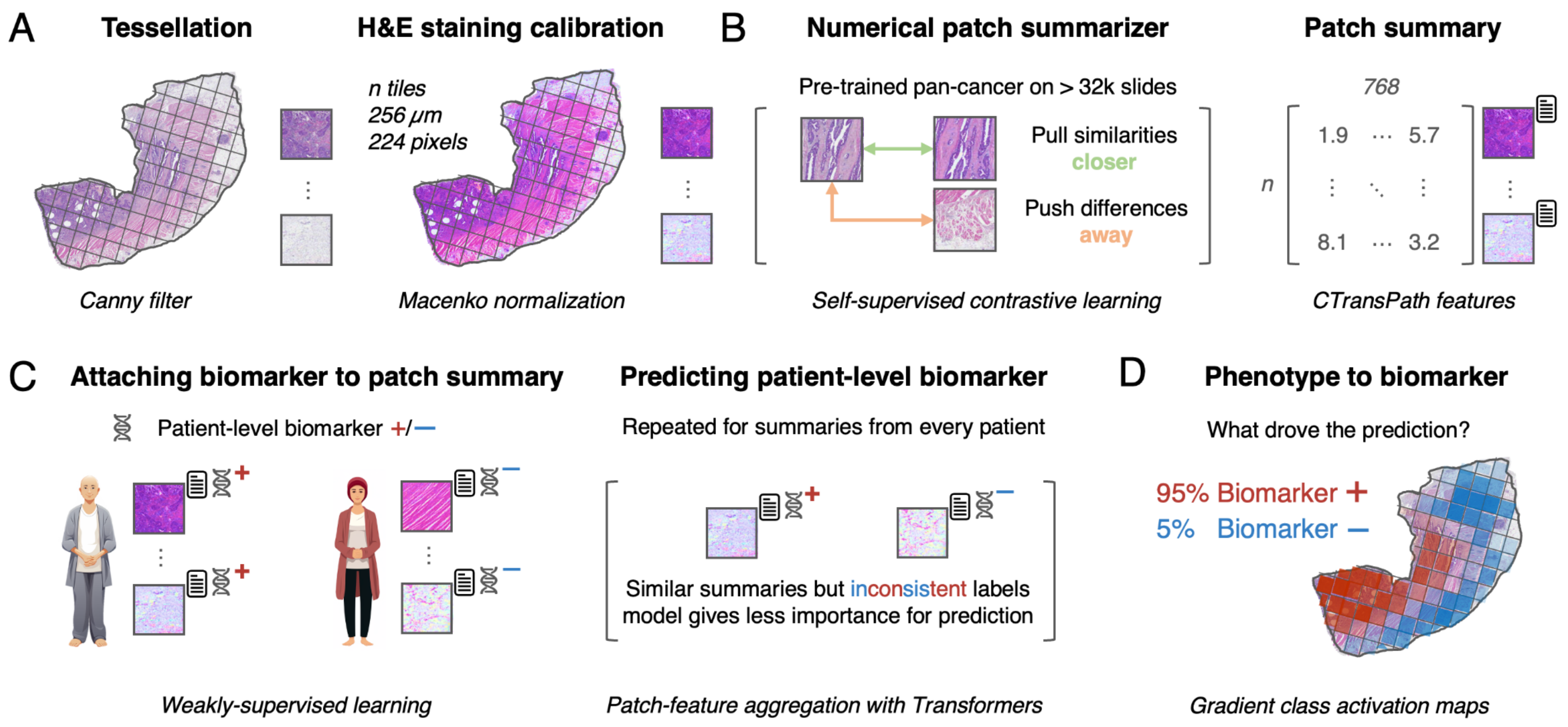

**Figure 2: Computational workflow from whole-slide image (WSI) to patient-level biomarker prediction.** **a.** Image preprocessing including WSI tessellation and hematoxylin and eosin (H&E)-staining color normalization. **b.** Feature extraction using self-supervised contrastive learning (CTransPath) to obtain feature vector representations for each tile in the WSI. **c.** Feature vectors and the target biomarker are input to the weakly-supervised model, which aggregates tile feature representation using a Transformer to provide a slide-level prediction. **d.** Explainability of the final predictions using gradient class activation maps for model interpretability purposes. The heatmaps highlight the most relevant areas for the model's decision-making for a biomarker prediction. The red and blue percentages are examples of the model's predicted likelihood of biomarker positivity or negativity.

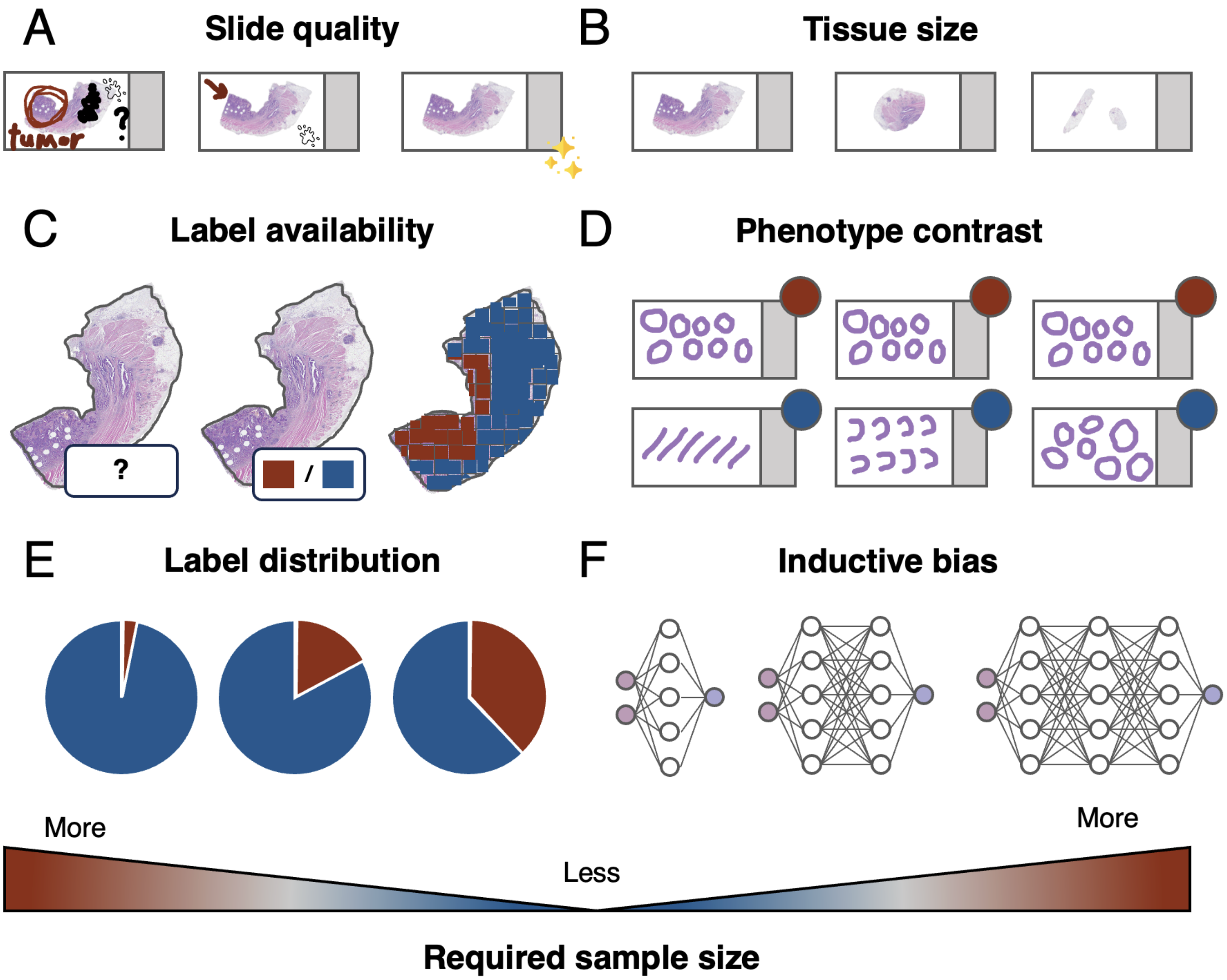

**Figure 3: Factors influencing the required sample size in computational pathology projects. a.** A slide with pen marks on the tissue and a dirty glass surface requires a higher sample size for modeling than a clean slide. **b.** Larger tissue slides, such as resections, contain more information than a biopsy for deep learning to find patterns. Therefore, large resections require less samples to model than small biopsies. **c.** Strongly supervised learning, having patch-level labels, requires less data than weakly- and self-supervised learning. **d.** A larger contrast between phenotypes of the positive and negative instance of a biomarker is easier to model, and therefore requires less samples than having instances with a similar phenotype. **e.** Rare positive instances of a biomarker, thus having an imbalanced dataset, require more samples for modeling than having a balanced distribution of the labels or annotations. **f.** Larger, more complex models like Transformers tend to have less inductive bias, making them adaptable to various tasks. Less inductive bias requires substantially larger amounts of data to model. In contrast, smaller models, while simpler and more biased, tend to generalize well on smaller sample sizes due to their more constrained nature.

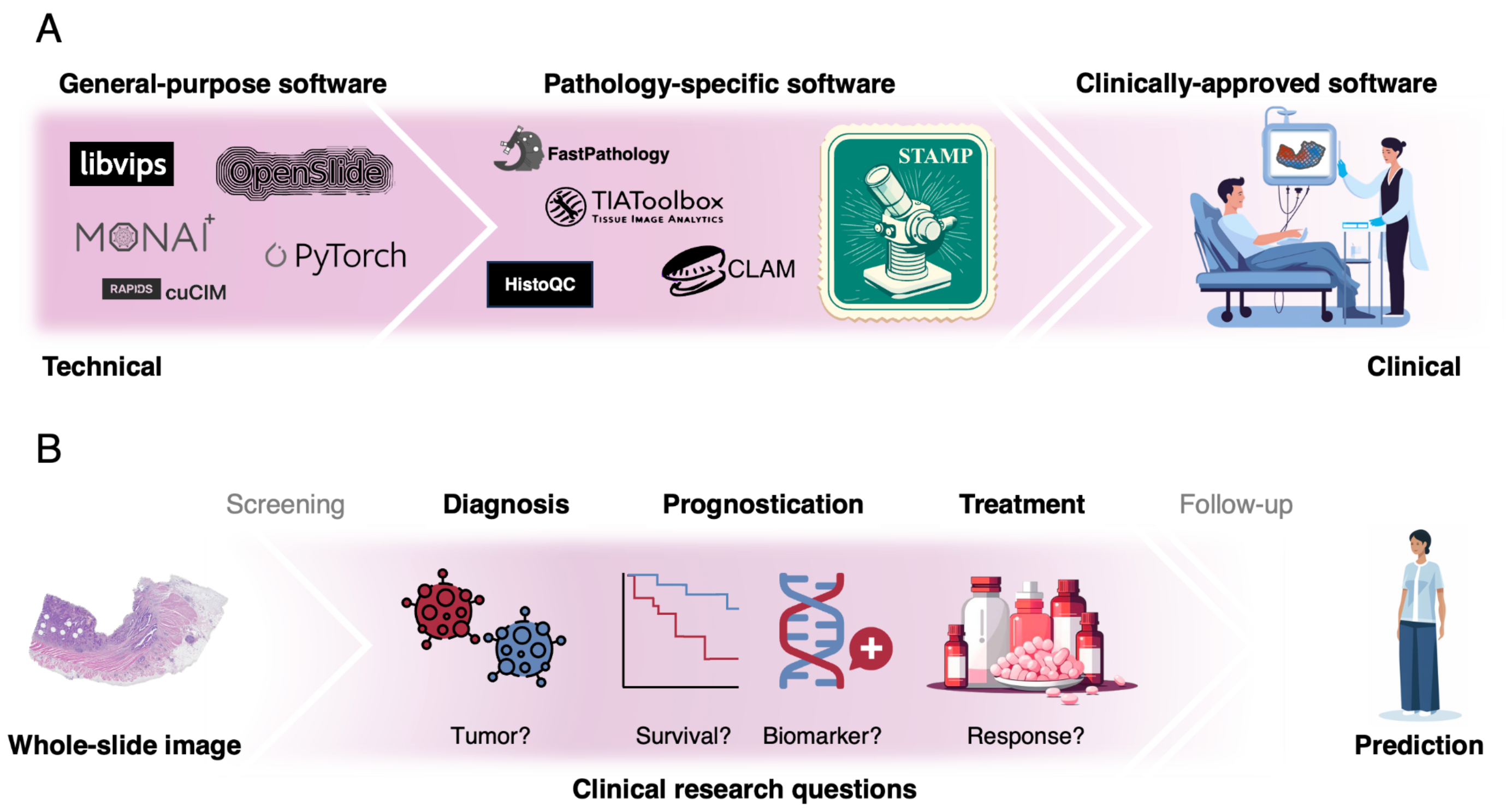

**Figure 4: Positioning of the STAMP software. a.** The STAMP protocol provides pathology-specific software which has been reduced to technical essence in order to provide state-of-the-art computational pathology methods, developed for usage by a clinical user-base. **b.** The STAMP protocol enables the answering of computational pathology research questions in the intermediate phases of the cancer patient journey, focusing on use-cases such as diagnosis, prognostication and treatment response prediction in a weakly-supervised manner.

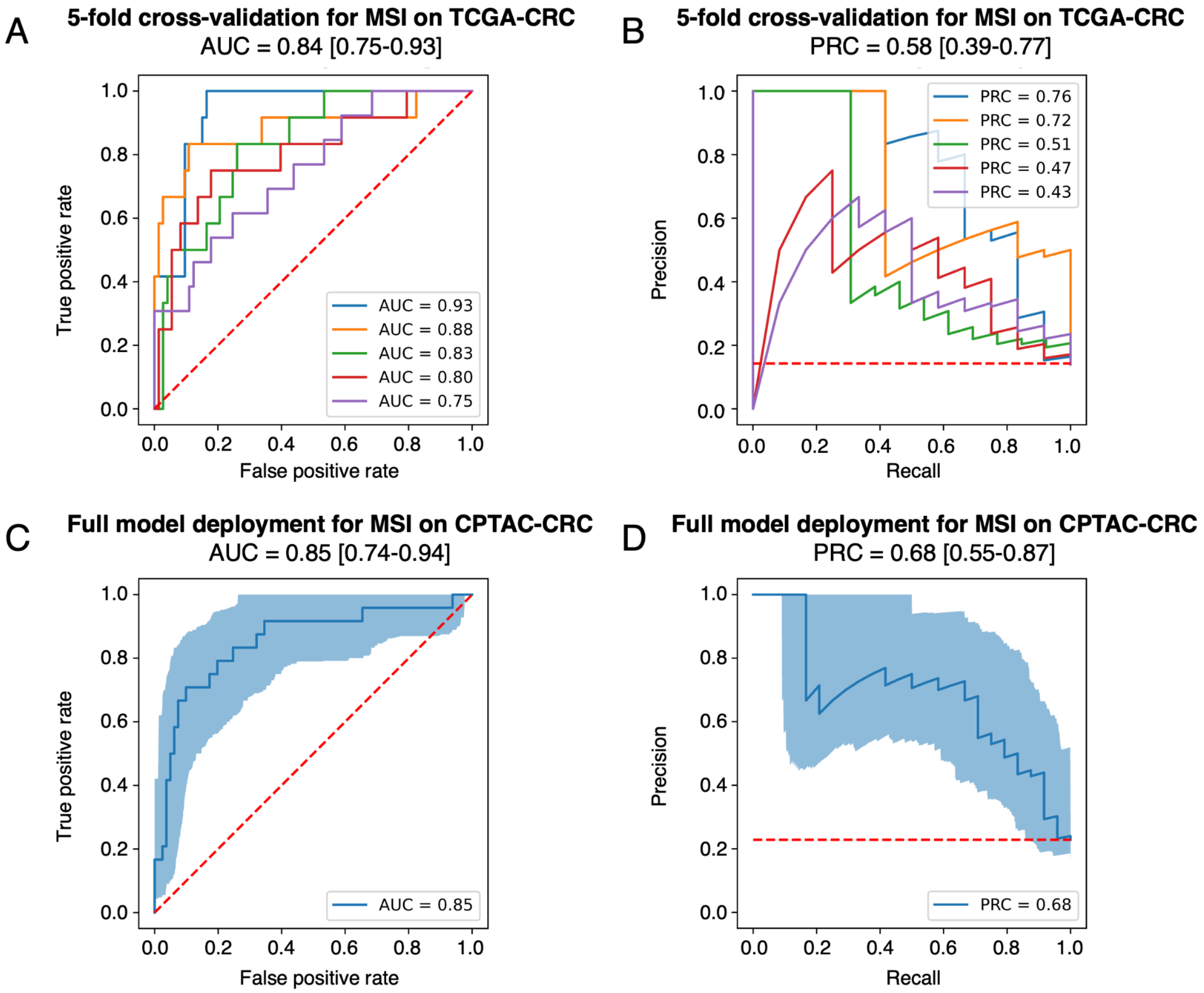

**Figure 5: Anticipated results of the evaluation phase of the protocol for the analysis of colorectal cancer (CRC) from The Cancer Genome Atlas (TCGA) and the Clinical Proteomic Tumor Analysis Consortium (CPTAC). a.** Area under the receiver-operator characteristic (AUROC) and **b.** area under the precision-recall characteristic (AUPRC) curves to evaluate the performance of 5-fold cross-validation for microsatellite instability high (MSI-H) trained on TCGA-CRC. **c.** AUROC and **d.** AUPRC curves to evaluate the performance of the final model trained and deployed on the external validation cohort CPTAC-CRC. The bootstrapped 95% confidence interval on the external dataset is depicted in light blue. The dotted red line signifies the performance of a random prediction model.

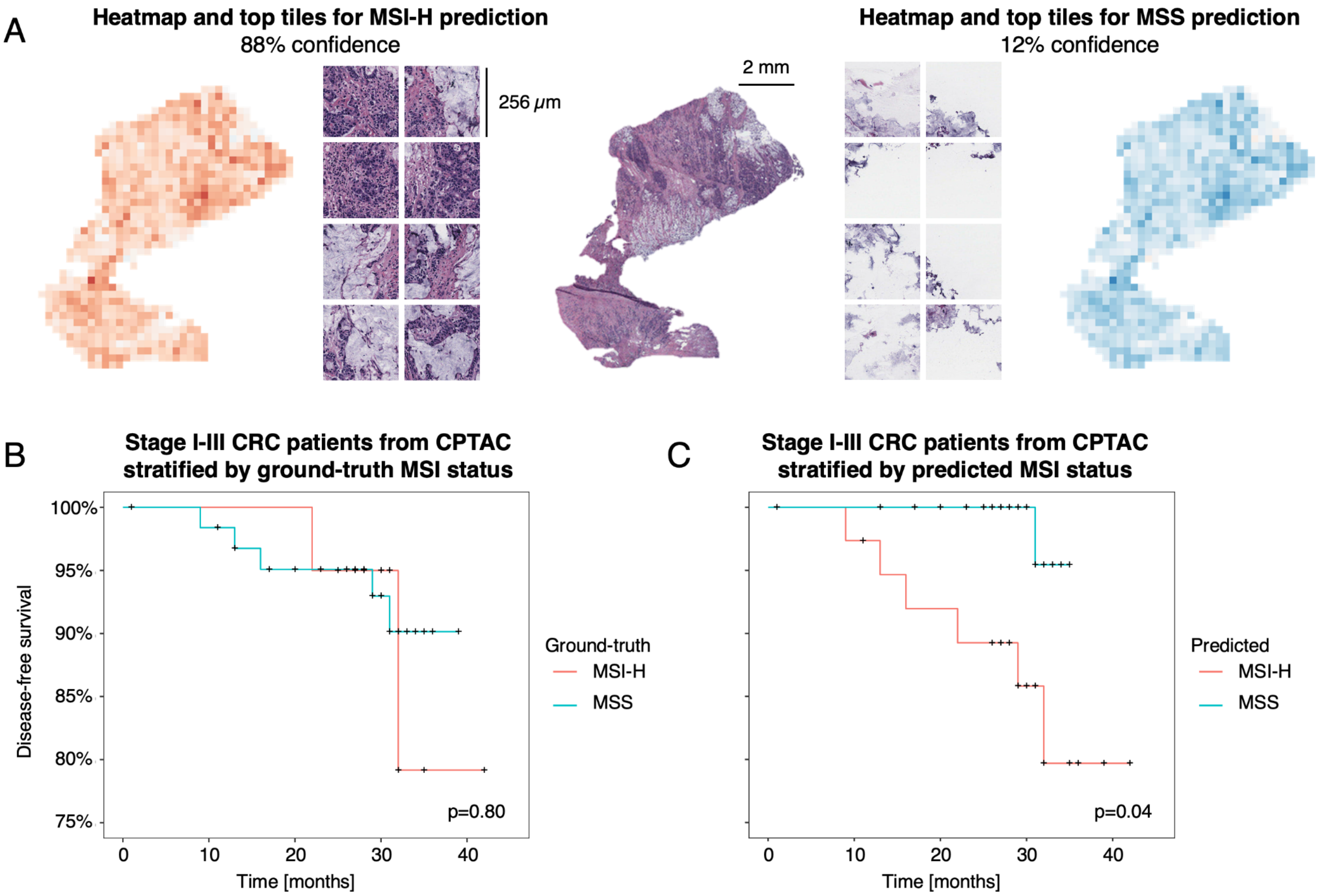

**Figure 6: Anticipated results of the translation phase of the protocol for the analysis of colorectal cancer (CRC) from the Clinical Proteomic Tumor Analysis Consortium (CPTAC). a.** The heatmap and top tiles for CPTAC slide 20CO003-2d266d2d-4bb3-436c-a3e7-06acb2, labeled as microsatellite instability high (MSI-H) for the ground-truth. Gradient class activation maps (gradCAM) are used to highlight the top influential tiles in the model's decision-making for either predicting MSI-H or microsatellite stable (MSS). In a binary prediction setting, the heatmaps are inversely correlated: the more a tile is positively attributing to MSI-H, the more it is negatively attributing to MSS. **b.** and **c.** show a Kaplan-Meier curve for stage I, II and III CRC patients from the CPTAC cohort. Stage IV patients are not analyzed separately due to only 1 datapoint with disease-free survival being available in the CPTAC cohort.

# Tables

**Table 1: Preprocessing configuration description in config.yaml file.**

| Argument | Example input | Description |
|---|---|---|
| output_dir | /home/storage/output_features | Path to save resulting feature vectors to. |
| wsi_dir | /home/storage/wsi_directory | Path to where the WSIs are stored. |
| cache_dir | /home/storage/cache_directory | Directory to save resulting intermediate preprocessing products to. |
| microns | 256 | Edge length of the tiles in microns. This is used for the microns-per-pixel (MPP) ratio to determine the slide resolution. The default MPP is 256/224 ≈ 1.14 which corresponds to ~9x magnification. |
| norm | true | Whether to perform stain normalization. |
| normalization_template | /home/STAMP/resources/normalization_template.jpg | Path to the normalization template used for stain normalization. |
| model_path | /home/STAMP/setup/ctranspath.pth | Path to the feature extraction model. |
| del_slide | false | Whether to remove the WSIs after preprocessing. |
| only_feature_extraction | false | Whether to only perform feature extraction. Only valid when intermediate products of prior preprocessing are available and are designated as the WSIs directory for only feature extraction. |
| cores | 8 | Amount of CPU cores to use. |
| device | cuda:0 | Select a device, GPU or CPU, to use for the preprocessing. |

**Table 2: Modeling configuration description in config.yaml file.**

| Argument | Example input | Description |
|---|---|---|
| clini_table | /home/storage/clinical_table.xlsx | Path to the table (.xlsx or .csv) containing the clinical parameters. The column with patient identifiers must be called PATIENTS. The column containing the biomarker to predict can be named anything without special characters. |
| slide_table | /home/storage/slide_table.csv | Path to the table (.xlsx or .csv) linking the patient identifiers to the filenames. The column with patient identifiers must be called |
| | | PATIENTS. The column with filenames must be called FILENAMES, and the entries should contain no file extensions. |
| feature_dir | /home/storage/output_features/STAMP_macenko_xiyuewang-ctranspath-7c998680 | Path to the directory containing the extracted feature files. |
| output_dir | /home/storage/modeling_output | Path to save modeling results to. |
| target_label | isMSIH | Name of the biomarker to predict. Should be the same name as present in the clinical table file. |
| categories | [MSI-H, MSS] | Optional indication of the categories, are automatically inferred otherwise. Contains the categories of the target label to predict. |
| cat_labels | [STAGE, SEX] | Optional addition of categorical tabular biomarkers for multimodal modeling. |
| cont_labels | [AGE] | Optional addition of continuous tabular biomarkers for multimodal modeling. |
| n_splits | 5 | Only applicable to cross-validation. The amount of folds with which the training cohort is split. Five folds yield a 5x 80/20 split for training/testing. |
| model_path | /home/storage/modeling_output/export.pkl | Only applicable to model deployment. The path to the model to deploy on an external cohort. |
| deploy_feature_dir | /home/storage/output_features_external/STAMP_macenko_xiyuewang-ctranspath-7c998680 | Only applicable to model deployment. The path to the features belonging to the external cohort to deploy a model on. |

**Table 3: Statistics configuration description in config.yaml file.**

| Argument | Example input | Description |
|---|---|---|
| pred_csvs | - /home/storage/modeling_output/fold-0/patient-preds.csv<br>- /home/storage/modeling_output/fold-1/patient-preds.csv<br>- /home/storage/modeling_output/fold-2/patient-preds.csv<br>- /home/storage/modeling_output/fold-3/patient-preds.csv<br>- /home/storage/modelin | Paths to the .csv files containing the patient predictions on the test set. Only needs 1 path for a single predictions file when evaluating a fully trained model on an external cohort. |
| | g_output/fold-4/patient-preds.csv | |
| target_label | isMSIH | Name of the biomarker to predict. Should be the same name as the biomarker that was trained on. |
| true_class | MSI-H | The category of the predicted biomarker considered to be the positive case. |
| output_dir | /home/storage/modeling_output | Path to save statistics results to. |

**Table 4: Heatmaps configuration description in config.yaml file.**

| Argument | Example input | Description |
|---|---|---|
| slide_name | 20CO003* | Slide filename without extension to create a heatmap for. A wildcard (*) can be used, with which multiple slide heatmaps following a similar naming convention are processed at once. |
| feature_dir | /home/storage/output_features_external/STAMP_macenko_xiyuewang-ctranspath-7c998680 | Name of the biomarker to predict. Should be the same name as the biomarker that was trained on. |
| wsi_dir | /home/storage/wsi_directory | Path to where the WSIs are stored. |
| model_path | /home/storage/modeling_output/export.pkl | Path to the trained model to generate heatmaps for. |
| output_dir | /home/storage/modeling_output/ | Path to save heatmaps results to. |
| n_toptiles | 8 | Number of tiles to be generated. Top tiles are the top influencers for the model's prediction of the biomarker categories the model is trained on. |

**Table 5: Common errors when running the STAMP software.**

| Step | Problem | Possible reason | Solution |
|---|---|---|---|
| 9 | Unsupported format error | The whole-slide image has a file extension which is not supported by OpenSlide. | Rescanning the slides into .tiff, or converting file to .ome.tiff using conversion tools (https://github.com/ome/bioformats). |
| 9 | Slide skipped due to missing resolution information | The metadata of the whole-slide image does not contain information on the resolution, expressed as microns-per-pixel. | Rescan the slide ensuring resolution info to be present, otherwise the slide cannot be processed and should be skipped. Estimating resolution is considered bad practice. |
| 9 | Not a JPEG file: starts with 0x00 | The whole-slide image has been corrupted during the scanning or | Re-download the file and try again. Otherwise the scan needs to be redone. |
| | 0x00 | downloading phase. | |
| 9 | CUDA error: invalid device ordinal | The GPU identifier selected for preprocessing is not available on the system. | Run the command nvidia-smi on your computer to identify how many GPUs there are, and choose a number (i.e., cuda:0) as input to the device argument of the preprocessing section. |
| 15 | The least populated class in y has only 1 member, which is too few. The minimum number of groups for any class cannot be less than 2. | One or more of the categories of the biomarker have only one occurrence. | Ensure that you manually specify the categories in the categories argument of the modeling section, so that no table artifacts ("None", "NA", or "-") are counted as classes. If the issue is not caused by artifacts, r\epeat step 5 to transform the data in a way that the categories are better distributed (>1 occurrence). |
| 15 | Value error: n_splits=5 cannot be greater than the number of members in each class. | There are more data splits requested than each category of the biomarker has samples. | Reduce the number of splits through the n_splits argument. In reality, it is recommended to gather more data, as no accurate model can be developed having such little amount of samples for each category. |
| 15, 17, 19 | No features found in feature_dir | The path for feature_dir in the modeling section is not correct. | Manually check the feature directory if there are .h5 files present with names that are put in the slide table (without file extensions) |
| 15, 17, 19 | Key error: ['name'] | Either the PATIENT, FILENAME or biomarker name has been misspelled or is not available in the clinical table. | The biomarker name in target_label needs to be exactly the same as how it is defined in the clinical table. The key error will indicate which name inconsistency it is complaining about. |
| 15, 19 | Key error: '[number] not in index' | The patient splits from a previous run were saved (folds.pt in the output_dir of the modeling section), and used again in the current run with different settings - those patients might not be available anymore. | Remove the output of previous runs which failed or were abruptly stopped, so that a new folds.pt file is generated. |
| 15, 19 | Value error: This DataLoader does not contain any batches | The data set is too small for the indicated batch size, which is 64. Sample sizes smaller than 64 cannot make a single batch to be run through the model. | It is recommended to gather more data, as no accurate model can be developed having such a small amount of samples for weakly-supervised deep learning problems in computational pathology. The batch size, however, can be changed manually in the code to |
| | | | accommodate smaller datasets, but is out of scope for this protocol. |
| 9, 15, 16, 17, 19, 20, 21 | `Missing required configuration keys: ['section.argument']` | The configuration file is incomplete, required arguments are missing. | Fill in the required arguments in the section that is indicated in the error message. |
| 9, 15, 16, 17, 19, 20, 21 | `Config file not found` | The configuration file is missing. | When running the CLI command to run one of the steps, use the --config flag to specify an absolute path to the configuration file. It is best practice to have the config.yaml file stored in the location where it was initially downloaded from the GitHub repository. |

**Table 6: Time estimates for hands-on execution and total duration of the steps in the STAMP protocol.**

| | **Lead** | **Hands-on execution** | **Total duration** |
|---|---|---|---|
| Steps 1-7, formal problem definition | Clinician | 1-2 weeks | 4-12 weeks |
| Steps 8-11, data preprocessing | Engineer | 2-6 hours | 1-6 weeks |
| Steps 12-18, modeling | Engineer | 3-9 hours | 1-3 weeks |
| Steps 19-23, evaluation | Engineer | 1-2 hours | 1-7 days |
| Steps 24-26, translation | Clinician | 1-2 weeks | 2-6 weeks |

# Boxes

**Box 1: Glossary for commonly used terms for communication between clinicians and engineers.**

| | |
|---|---|
| Argument | Information provided to a software program to modify its behavior or control its functioning. The arguments are provided in the configuration file to inform the STAMP software where data is located for processing. |
| Architecture | The design and structure for a deep learning algorithm, detailing how its components are organized and function together. An example of an architecture is the Transformer. |
| AUROC | Area under the receiver operator characteristic. A statistical measure in classification problems indicating the model's ability to distinguish between different predicted labels. |
| AUPRC | Area under the precision-recall characteristic. A statistical measure showing model performance in classification problems, particularly useful with imbalanced data. |
| Clinical table | A structured table containing various information or labels corresponding to individual patients, in CSV or XLSX format. |
| Command line interface | A text-based method to communicate and interact with the STAMP software and computer programs in general. Also referred to as the terminal. It is used to interact with the STAMP software. |
| Downstream task | Additional tasks or processes connected to the model's predictions for further analysis or decision-making. The downstream task refers to the modeling procedure to link the extracted features to the biomarker. |
| Epoch | One iteration of the model through the entire training dataset during the learning process. The amount of epochs determine how often the dataset is seen during training. In theory, the more epochs used for training, the more exploration room the model has to explore patterns in the data. |
| Feature extraction | The process of obtaining essential information from raw data (image patches) to create condensed, lower-dimensional representations (features). In short, information is summarized to create a higher level concept of the information present in the image patches. |
| Feature | A vector of numbers depicting a condensed representation of image patches. Due to the black-box nature of deep learning, the meaning of these numbers are unknown. |
| Heatmaps | Visual representations indicating the most influential regions in data that the model used for its decisions. Specifically, gradient class-activation maps[67] are used to highlight the key influential regions. |
| Inference | Using a trained model to make predictions by providing new, unseen data as input. Inference, also referred to as deploying a model, is a static procedure, meaning the model's properties are not altered due to new data being shown. |
| Label | Categories within the target biomarker, such as deceased/alive for a survival target. |
| Model | A deep learning architecture that has been trained on data, and enables prediction of new data inputs based on patterns it has previously learned. |
| Patient identifier | A unique label or code assigned to each patient within the dataset. |
| Prediction scores | Numeric values generated by the model indicating the likelihood of a particular prediction or classification. |
| Slide table | A table that links files to specific patients, in CSV or XLSX format. |
| Target | The biomarker the model is trained on to predict, such as survival. |
| Top tiles | Image patches that substantially influence the model's predictions, ranked by their magnitude of patch-wise classification score. |

**Box 2: Installation instructions for the STAMP software.**

**Installation steps:**

1. **Install OpenSlide**: Use the following command or refer to the official installation instructions (https://openslide.org/download/#distribution-packages):

   ```
   apt update && apt install -y openslide-tools libgl1-mesa-glx
   ```

2. **Install Conda**: Follow the steps outlined in the Conda installation guide found at https://conda.io/projects/conda/en/latest/user-guide/install/index.html to set up Conda on your local machine. Create an environment with Python 3.10 and activate it:

   ```
   conda create -n stamp python=3.10
   conda activate stamp
   conda install -c conda-forge libstdcxx-ng=12
   ```

3. **Install the STAMP package**: Use pip to install the STAMP package from the GitHub repository: https://github.com/KatherLab/STAMP.

   ```
   pip install git+https://github.com/KatherLab/STAMP
   ```

4. **Download required resources**: After installation, download essential resources like the CTransPath feature extractor's weights by running the command:

   ```
   stamp setup
   ```

Upon successful installation, the *stamp* command can be used in the command line interface.

**Box 3: Overview of all command line interface (CLI) commands of the STAMP software.**

| | |
|---|---|
| `stamp setup` | Download required resources |
| `stamp config` | Show the configuration settings |
| `stamp preprocess` | Preprocess WSIs and extract features |
| `stamp crossval` | Train n_splits models using cross-validation |
| `stamp train` | Train a single model on the entire training cohort |
| `stamp deploy` | Deploy a trained model on an external testing cohort |
| `stamp statistics` | Compute the AUROC, AUPRC and corresponding 95%CI metrics |
| `stamp heatmaps` | Generate heatmaps and corresponding toptiles |

**Box 4: Example of slide table used in the STAMP protocol.**

The slide table is a file (*.xlsx* or *.csv*) which contains two columns, a *PATIENT* column, containing the patient identifiers as string, and the *FILENAME* column, containing the feature matrix names as string without the file extension (*.h5*). A patient identifier can be linked to multiple WSIs by repeating the patient identifier and adding the additional corresponding feature matrix names. In the downstream classification task, the features of all the slides belonging to a single patient will be concatenated into one big feature matrix during runtime. This is wished behavior for modeling a patient-level target in homogeneous tumors, such as a genetic alteration, which might contain valuable information in all the observed slides. However, when modeling a slide-level target, such as a pathologist-assigned score, it is recommended to treat the WSIs as individual instances and not concatenate their features in the downstream classification task. In this case, each slide should have a unique patient identification label to avoid concatenation of slide features during runtime.

| **PATIENT** | **FILENAME** |
|---|---|
| ID_1337 | slide_ID_1337_1 |
| ID_1337 | slide_ID_1337_2 |
| ID_1999 | slide_ID_1999_1 |
| ID_1608 | slide_ID_1608_1 |

**Box 5: Example of clinical table used in the STAMP protocol.**

The clinical table is a file (*.xlsx* or *.csv*) which contains at least two columns, one column with the patient identifiers, *PATIENT*, and a column with the biomarker data to train the model on, which is a string with any chosen name without special characters. The clinical table can have other columns, as long as each column name is a unique string. The absolute path of the clinical table is used as input for the *clini_table* argument in the modeling section of the configuration file. To specify the biomarker to be modeled, provide the name of the corresponding column name in the clinical table as a string input for the *target_label* argument. The protocol supports multimodality by adding categorical and continuous tabular variables to the extracted feature vector from the image[32], using the *cat_variable* and *cont_variable* arguments, respectively. Except for the *PATIENT* column, the column names in the clinical table that are not explicitly provided as an argument are not used for modeling. It is recommended to define the desired categories to train on using the *categories* argument in the *modeling* section. Missing values in the clinical table should be empty cells, because strings or characters, like "*NA*", "*NaN*", "*None*", or "-", will be seen as categories, unless the categories are explicitly defined using the *categories* argument.

| **PATIENT** | **isMSIH** |
|---|---|
| ID_1337 | MSI-H |
| ID_1999 | MSS |
| ID_1608 | |
| ID_0311 | MSI-H |

**Box 6: Data splitting mechanisms used in the STAMP protocol.**

There are four main data components that are required for proper development of a model to predict biomarkers directly from WSI: training data, validation data, test data and external test data. Training and validation data are seen by the model during training time, which is then evaluated using the unseen test data. Aforementioned data splits are usually from the same patient cohort. To test generalizability of the model, its performance is measured on the unseen external test set. By default, 5-fold cross-validation is performed on the cohort, splitting the provided clinical table from **step** 12 into 5 permutations of an 80-20 split for modeling (using training and validation data) and testing data. Consequently, 64% of the provided cohort is used for training, 16% for validation, and 20% for testing, repeated for 5 iterations, stratified by the target variable (**Suppl. Fig. 1**). After 5-fold cross-validation, another training run is performed on the entire cohort to yield a single model. Again, the data is split 80-20, but this time solely for training and validation, as the external test set will be used to measure performance instead. This data splitting mechanism has been chosen due to its consistent usage for modeling in computational pathology in previous studies[26,68,69]. The external test set is unused and unseen during model training to avoid data leakage. If no external test set is available, it is recommended to create pseudo-external test sets from the training cohort, ensuring patients' slides and hospital of origin to be mutually exclusive in the training and testing cohort to reduce confounding factors during the modeling[56].

# References

1. Shmatko, A., Ghaffari Laleh, N., Gerstung, M. & Kather, J. N. Artificial intelligence in histopathology: enhancing cancer research and clinical oncology. *Nat Cancer* **3**, 1026–1038 (2022).
2. Ghaffari Laleh, N. *et al.* Benchmarking weakly-supervised deep learning pipelines for whole slide classification in computational pathology. *Med. Image Anal.* **79**, 102474 (2022).
3. Foersch, S. *et al.* Deep learning for diagnosis and survival prediction in soft tissue sarcoma. *Ann. Oncol.* **32**, 1178–1187 (2021).
4. Klein, C. *et al.* Artificial intelligence for solid tumour diagnosis in digital pathology. *Br. J. Pharmacol.* **178**, 4291–4315 (2021).
5. Woerl, A.-C. *et al.* Deep Learning Predicts Molecular Subtype of Muscle-invasive Bladder Cancer from Conventional Histopathological Slides. *Eur. Urol.* **78**, 256–264 (2020).
6. Hong, R., Liu, W., DeLair, D., Razavian, N. & Fenyö, D. Predicting endometrial cancer subtypes and molecular features from histopathology images using multi-resolution deep learning models. *Cell Rep Med* **2**, 100400 (2021).
7. Kather, J. N. *et al.* Predicting survival from colorectal cancer histology slides using deep learning: A retrospective multicenter study. *PLoS Med.* **16**, e1002730 (2019).
8. Laleh, N. G., Echle, A., Muti, H. S. & Hewitt, K. J. Deep Learning for interpretable end-to-end survival (E-ESurv) prediction in gastrointestinal cancer histopathology. *MICCAI Workshop* (2021).
9. Foersch, S. *et al.* Multistain deep learning for prediction of prognosis and therapy response in colorectal cancer. *Nat. Med.* **29**, 430–439 (2023).
10. Wang, C.-W. *et al.* Weakly supervised deep learning for prediction of treatment effectiveness on ovarian cancer from histopathology images. *Comput. Med. Imaging Graph.* **99**, 102093 (2022).
11. Ghaffari Laleh, N., Ligero, M., Perez-Lopez, R. & Kather, J. N. Facts and Hopes on the Use of Artificial Intelligence for Predictive Immunotherapy Biomarkers in Cancer. *Clin. Cancer Res.* **29**, 316–323 (2023).
12. Kather, J. N. *et al.* Pan-cancer image-based detection of clinically actionable genetic alterations. *Nat Cancer* **1**, 789–799 (2020).
13. Kanavati, F. *et al.* Weakly-supervised learning for lung carcinoma classification using deep learning. *Sci. Rep.* **10**, 9297 (2020).

14. Wang, X. *et al.* Weakly Supervised Deep Learning for Whole Slide Lung Cancer Image Analysis. *IEEE Trans Cybern* **50**, 3950–3962 (2020).

15. Kather, J. N. *et al.* Deep learning can predict microsatellite instability directly from histology in gastrointestinal cancer. *Nat. Med.* **25**, 1054–1056 (2019).

16. Bilal, M. *et al.* Development and validation of a weakly supervised deep learning framework to predict the status of molecular pathways and key mutations in colorectal cancer from routine histology images: a retrospective study. *Lancet Digit Health* **3**, e763–e772 (2021).

17. Schrammen, P. L. *et al.* Weakly supervised annotation-free cancer detection and prediction of genotype in routine histopathology. *J. Pathol.* **256**, 50–60 (2022).

18. Echle, A. *et al.* Clinical-Grade Detection of Microsatellite Instability in Colorectal Tumors by Deep Learning. *Gastroenterology* **159**, 1406–1416.e11 (2020).

19. Zeng, Q. *et al.* Artificial intelligence predicts immune and inflammatory gene signatures directly from hepatocellular carcinoma histology. *J. Hepatol.* **77**, 116–127 (2022).

20. Jaroensri, R. *et al.* Deep learning models for histologic grading of breast cancer and association with disease prognosis. *NPJ Breast Cancer* **8**, 113 (2022).

21. Li, C. *et al.* Weakly supervised mitosis detection in breast histopathology images using concentric loss. *Med. Image Anal.* **53**, 165–178 (2019).

22. Zheng, Q. *et al.* A Weakly Supervised Deep Learning Model and Human-Machine Fusion for Accurate Grading of Renal Cell Carcinoma from Histopathology Slides. *Cancers* **15**, (2023).

23. Muti, H. S. *et al.* *The Aachen Protocol for Deep Learning Histopathology: A hands-on guide for data preprocessing*. doi:10.5281/zenodo.3694994.

24. Graziani, M. *et al.* Attention-Based Interpretable Regression of Gene Expression in Histology. in *Interpretability of Machine Intelligence in Medical Image Computing: 5th International Workshop, iMIMIC 2022, Held in Conjunction with MICCAI 2022, Singapore, Singapore, September 22, 2022, Proceedings* 44–60 (Springer-Verlag, 2022).

25. Campanella, G. *et al.* Clinical-grade computational pathology using weakly supervised deep learning on whole slide images. *Nat. Med.* **25**, 1301–1309 (2019).

26. Schmauch, B. *et al.* A deep learning model to predict RNA-Seq expression of tumours from whole slide images. *Nat. Commun.* **11**, 1–15 (2020).

27. Lu, M. Y. *et al.* AI-based pathology predicts origins for cancers of unknown primary. *Nature* **594**, 106–110 (2021).
28. Wagner, S. J. *et al.* Built to Last? Reproducibility and Reusability of Deep Learning Algorithms in Computational Pathology. *Mod. Pathol.* **37**, 100350 (2023).
29. Veldhuizen, G. P. *et al.* Deep learning-based subtyping of gastric cancer histology predicts clinical outcome: a multi-institutional retrospective study. *Gastric Cancer* **26**, 708–720 (2023).
30. Muti, H. S. *et al.* Deep learning trained on lymph node status predicts outcome from gastric cancer histopathology: a retrospective multicentric study. *Eur. J. Cancer* 113335 (2023).
31. Saldanha, O. L. *et al.* Direct prediction of genetic aberrations from pathology images in gastric cancer with swarm learning. *Gastric Cancer* **26**, 264–274 (2023).
32. Niehues, J. M. *et al.* Generalizable biomarker prediction from cancer pathology slides with self-supervised deep learning: A retrospective multi-centric study. *Cell Rep Med* 100980 (2023).
33. Wagner, S. J. *et al.* Transformer-based biomarker prediction from colorectal cancer histology: A large-scale multicentric study. *Cancer Cell* **41**, 1650–1661.e4 (2023).
34. Chatterji, S. *et al.* Prediction models for hormone receptor status in female breast cancer do not extend to males: further evidence of sex-based disparity in breast cancer. *NPJ Breast Cancer* **9**, 91 (2023).
35. Hewitt, K. J. *et al.* Direct image to subtype prediction for brain tumors using deep learning. *Neurooncol. Adv.* (2023) doi:10.1093/noajnl/vdad139.
36. Saldanha, O. L. *et al.* Self-supervised attention-based deep learning for pan-cancer mutation prediction from histopathology. *NPJ Precis Oncol* **7**, 35 (2023).
37. Loeffler, C. M. L. *et al.* Direct prediction of Homologous Recombination Deficiency from routine histology in ten different tumor types with attention-based Multiple Instance Learning: a development and validation study. *medRxiv* 2023.03.08.23286975 (2023) doi:10.1101/2023.03.08.23286975.
38. El Nahhas, O. S. M. *et al.* Regression-based Deep-Learning predicts molecular biomarkers from pathology slides. *arXiv [cs.CV]* (2023).
39. Wang, X. *et al.* Transformer-based unsupervised contrastive learning for histopathological image classification. *Med. Image Anal.* **81**, 102559 (2022).
40. Greenson, J. K. *et al.* Pathologic predictors of microsatellite instability in colorectal cancer. *Am. J. Surg. Pathol.* **33**, 126–133 (2009).

41. Causality in digital medicine. *Nat. Commun.* **12**, 5471 (2021).

42. Cancer Genome Atlas Network. Comprehensive molecular characterization of human colon and rectal cancer. *Nature* **487**, 330–337 (2012).

43. Ellis, M. J. *et al.* Connecting genomic alterations to cancer biology with proteomics: the NCI Clinical Proteomic Tumor Analysis Consortium. *Cancer Discov.* **3**, 1108–1112 (2013).

44. Goode, A., Gilbert, B., Harkes, J., Jukic, D. & Satyanarayanan, M. OpenSlide: A vendor-neutral software foundation for digital pathology. *J. Pathol. Inform.* **4**, 27 (2013).

45. Bankhead, P. *et al.* QuPath: Open source software for digital pathology image analysis. *Sci. Rep.* **7**, 1–7 (2017).

46. Paszke, A. *et al.* PyTorch: An Imperative Style, High-Performance Deep Learning Library. *arXiv [cs.LG]* (2019).

47. Jorge Cardoso, M. *et al.* MONAI: An open-source framework for deep learning in healthcare. Preprint at https://arxiv.org/abs/2211.02701 (2022).

48. Martinez, K. & Cupitt, J. VIPS - a highly tuned image processing software architecture. in *IEEE International Conference on Image Processing 2005* (IEEE, 2005). doi:10.1109/icip.2005.1530120.

49. Janowczyk, A., Zuo, R., Gilmore, H., Feldman, M. & Madabhushi, A. HistoQC: An Open-Source Quality Control Tool for Digital Pathology Slides. *JCO Clin Cancer Inform* **3**, 1–7 (2019).

50. Pedersen, A. *et al.* FastPathology: An Open-Source Platform for Deep Learning-Based Research and Decision Support in Digital Pathology. *IEEE Access* **9**, 58216–58229 (2021).

51. Pocock, J. *et al.* TIAToolbox as an end-to-end library for advanced tissue image analytics. *Commun. Med.* **2**, 120 (2022).

52. Lu, M. Y. *et al.* Data-efficient and weakly supervised computational pathology on whole-slide images. *Nat Biomed Eng* **5**, 555–570 (2021).

53. Verghese, G. *et al.* Computational pathology in cancer diagnosis, prognosis, and prediction - present day and prospects. *J. Pathol.* **260**, 551–563 (2023).

54. Saillard, C. *et al.* Validation of MSIntuit as an AI-based pre-screening tool for MSI detection from colorectal cancer histology slides. *Nat. Commun.* **14**, 6695 (2023).

55. Macenko, M. *et al.* A method for normalizing histology slides for quantitative analysis. in *2009 IEEE International Symposium on Biomedical Imaging: From Nano to Macro* 1107–1110 (2009).

56. Howard, F. M. *et al.* The impact of site-specific digital histology signatures on deep learning model accuracy and bias. *Nat. Commun.* **12**, 4423 (2021).

57. Canny, J. A Computational Approach to Edge Detection. *IEEE Trans. Pattern Anal. Mach. Intell.* **PAMI-8**, 679–698 (1986).

58. Wölflein, G. *et al.* A Good Feature Extractor Is All You Need for Weakly Supervised Learning in Histopathology. *arXiv [cs.CV]* (2023).

59. Comes, M. C. *et al.* A deep learning model based on whole slide images to predict disease-free survival in cutaneous melanoma patients. *Sci. Rep.* **12**, 20366 (2022).

60. Jiang, S., Suriawinata, A. A. & Hassanpour, S. MHAttnSurv: Multi-head attention for survival prediction using whole-slide pathology images. *Comput. Biol. Med.* **158**, 106883 (2023).

61. Sounderajah, V. *et al.* Developing a reporting guideline for artificial intelligence-centred diagnostic test accuracy studies: the STARD-AI protocol. *BMJ Open* **11**, e047709 (2021).

62. Collins, G. S. *et al.* Protocol for development of a reporting guideline (TRIPOD-AI) and risk of bias tool (PROBAST-AI) for diagnostic and prognostic prediction model studies based on artificial intelligence. *BMJ Open* **11**, e048008 (2021).

63. Trautmann, K. *et al.* Chromosomal instability in microsatellite-unstable and stable colon cancer. *Clin. Cancer Res.* **12**, 6379–6385 (2006).

64. Lin, E. I. *et al.* Mutational profiling of colorectal cancers with microsatellite instability. *Oncotarget* **6**, 42334–42344 (2015).

65. Boland, C. R. & Goel, A. Microsatellite instability in colorectal cancer. *Gastroenterology* **138**, 2073–2087.e3 (2010).

66. Battaglin, F., Naseem, M., Lenz, H.-J. & Salem, M. E. Microsatellite instability in colorectal cancer: overview of its clinical significance and novel perspectives. *Clin. Adv. Hematol. Oncol.* **16**, 735–745 (2018).

67. Selvaraju, R. R. *et al.* Grad-CAM: Visual Explanations from Deep Networks via Gradient-Based Localization. in *2017 IEEE International Conference on Computer Vision (ICCV)* 618–626 (IEEE, 2017).

68. Pataki, B. Á. *et al.* HunCRC: annotated pathological slides to enhance deep learning applications in colorectal cancer screening. *Scientific Data* **9**, 1–7 (2022).

69. Cheng, J. *et al.* Computational analysis of pathological images enables a better diagnosis of TFE3

Xp11.2 translocation renal cell carcinoma. *Nat. Commun.* **11**, 1–9 (2020).

# Author contributions statements

# Data availability

Histopathology slides and genomics data from TCGA and CPTAC were used to train and validate the models. The slides for TCGA are available at https://portal.gdc.cancer.gov/. The slides for CPTAC are available at https://proteomics.cancer.gov/data-portal. The molecular and clinical data for TCGA and CPTAC used in the experiments are available at https://github.com/KatherLab/cancer-metadata.

# Code availability

The open-source STAMP software for the implementation of the MSI experiments is available on GitHub (https://github.com/KatherLab/STAMP).

## Additional information

As per the ICMJE guidelines of April 2023, we hereby disclose that the following tools were used to write this article. Microsoft Word and Google Documents as Word processing software, ChatGPT-4 for checking and correcting spelling and grammar, Midjourney V5.2 for figure icon generation.

### Ethics statement

We examined anonymized patient samples from several academic institutions in this investigation. CPTAC and TCGA did not require formal ethics approval for a retrospective study of anonymised samples. The overall analysis was approved by the Ethics commission of the Medical Faculty of the Technical University Dresden (BO-EK-444102022).

### Author contributions

OSMEN and JNK designed the protocol. OSMEN, MVT, GW and TL developed the software and wrote technical documentation. OSMEN, MVT, GW, TL and ML tested the software. OSMEN, JNK and KJH interpreted and analyzed the data. All authors wrote and reviewed the protocol and approved the final version for submission.

### Conflicts of Interest

OSMEN holds shares in StratifAI GmbH.

FK holds shares in StratifAI GmbH.

DT holds shares in StratifAI GmbH.

JNK declares consulting services for Owkin, France, DoMore Diagnostics, Norway, Panakeia, UK, Scailyte, Switzerland, Mindpeak, Germany, and Histofy, UK; furthermore he holds shares in StratifAI GmbH, Germany, and has received honoraria for lectures by AstraZeneca, Bayer, Eisai, MSD, BMS, Roche, Pfizer and Fresenius. DT received honoraria for lectures by Bayer and holds shares in StratifAI GmbH, Germany. SF has received honoraria from MSD and BMS.

## Funding

JNK is supported by the German Federal Ministry of Health (DEEP LIVER, ZMVI1-2520DAT111), the German Cancer Aid (DECADE, 70115166), the German Federal Ministry of Education and Research (PEARL, 01KD2104C; CAMINO, 01EO2101; SWAG, 01KD2215A; TRANSFORM LIVER, 031L0312A; TANGERINE, 01KT2302 through ERA-NET Transcan), the German Academic Exchange Service (SECAI, 57616814), the German Federal Joint Committee (TransplantKI, 01VSF21048) the European Union's Horizon Europe and innovation programme (ODELIA, 101057091; GENIAL, 101096312) and the National Institute for Health and Care Research (NIHR, NIHR213331) Leeds Biomedical Research Centre.
GW is supported by Lothian NHS.

DT is supported by the German Federal Ministry of Education and Research (SWAG, 01KD2215A; TRANSFORM LIVER), the European Union's Horizon Europe and innovation programme (ODELIA, 101057091).

SF is supported by the German Federal Ministry of Education and Research (SWAG, 01KD2215A), the German Cancer Aid (DECADE, 70115166) and the German Research Foundation (504101714).
SJW was supported by the Helmholtz Association under the joint research school "Munich School for Data Science - MUDS" and the Add-on Fellowship of the Joachim Herz Foundation.

The views expressed are those of the author(s) and not necessarily those of the NHS, the NIHR or the Department of Health and Social Care. This work was funded by the European Union. Views and opinions expressed are however those of the author(s) only and do not necessarily reflect those of the European Union. Neither the European Union nor the granting authority can be held responsible for them.

## Acknowledgements

Thank you to the testers of the protocol, Srividhya Sainath, Oliver Lester Saldanha, Laura Žigutytė, Cornelius Kummer, Garazi Serna, Kevin Boehm and Lawrence Shaktah who executed the STAMP protocol on various systems at cancer centers around the world.

# Supplementary Figures and Tables

**Suppl. Table 1: Overview of the main software packages of the protocol.**

| Library | Description | Website |
|---|---|---|
| `pandas` | A data manipulation and analysis library for Python. It provides data structures and data analysis tools. | pandas.pydata.org |
| `matplotlib` | A library for creating static, interactive, and animated visualizations in Python which works well with NumPy and Pandas data structures. | matplotlib.org |
| `scikit-learn` | A simple and efficient tool for data mining and data analysis. It provides various machine learning algorithms and tools for data processing. | scikit-learn.org |
| `tqdm` | A progress bar for loops and iterables in Python. It provides a way to track the progress of iterations. | github.com/tqdm/tqdm |
| `fastai` | A deep learning library built on top of PyTorch. It simplifies the process of applying deep learning techniques and supports computer vision, text, tabular data, and collaborative filtering. | docs.fast.ai |
| `torch` | PyTorch is an open-source machine learning framework that accelerates the path from research prototyping to production deployment. It provides tensors and dynamic neural networks. | pytorch.org |
| `torchvision` | A package that provides datasets, model architectures, and common image transformations for computer vision in PyTorch. | pytorch.org/vision/stable/index.html |
| `h5py` | A package that provides a Pythonic interface to the HDF5 binary data format. It allows the storage and manipulation of large amounts of numerical data efficiently. | www.h5py.org |
| `jaxtyping` | A package that provides additional typing annotations for JAX, which is a system for high-performance machine learning research. | github.com/google/jaxtyping |
| `einops` | A package for tensor manipulation, especially focusing on reshaping operations in a flexible and concise way. | github.com/arogozhnikov/einops |
| `omegaconf` | A package that provides a flexible configuration system in Python, primarily used in machine learning projects. | omegaconf.readthedocs.io |
| `openslide-python` | A Python interface to the OpenSlide library, which allows for reading whole-slide images in various formats used in digital pathology. | openslide.org/api/python/ |
| `opencv-python` | An open-source computer vision and machine learning software library. It provides tools and algorithms for image processing and computer vision tasks. | opencv.org |
| `numba` | A just-in-time compiler for Python that translates portions of Python code to machine code. It is used to accelerate numerical algorithms. | numba.pydata.org |
| `gdown` | A tool for downloading files from Google Drive. It simplifies the process of downloading large files programmatically. | github.com/wkentaro/gdown |
| `openpyxl` | A package for reading and writing Excel files in Python. It allows manipulation of Excel files using Python code. | openpyxl.readthedocs.io |

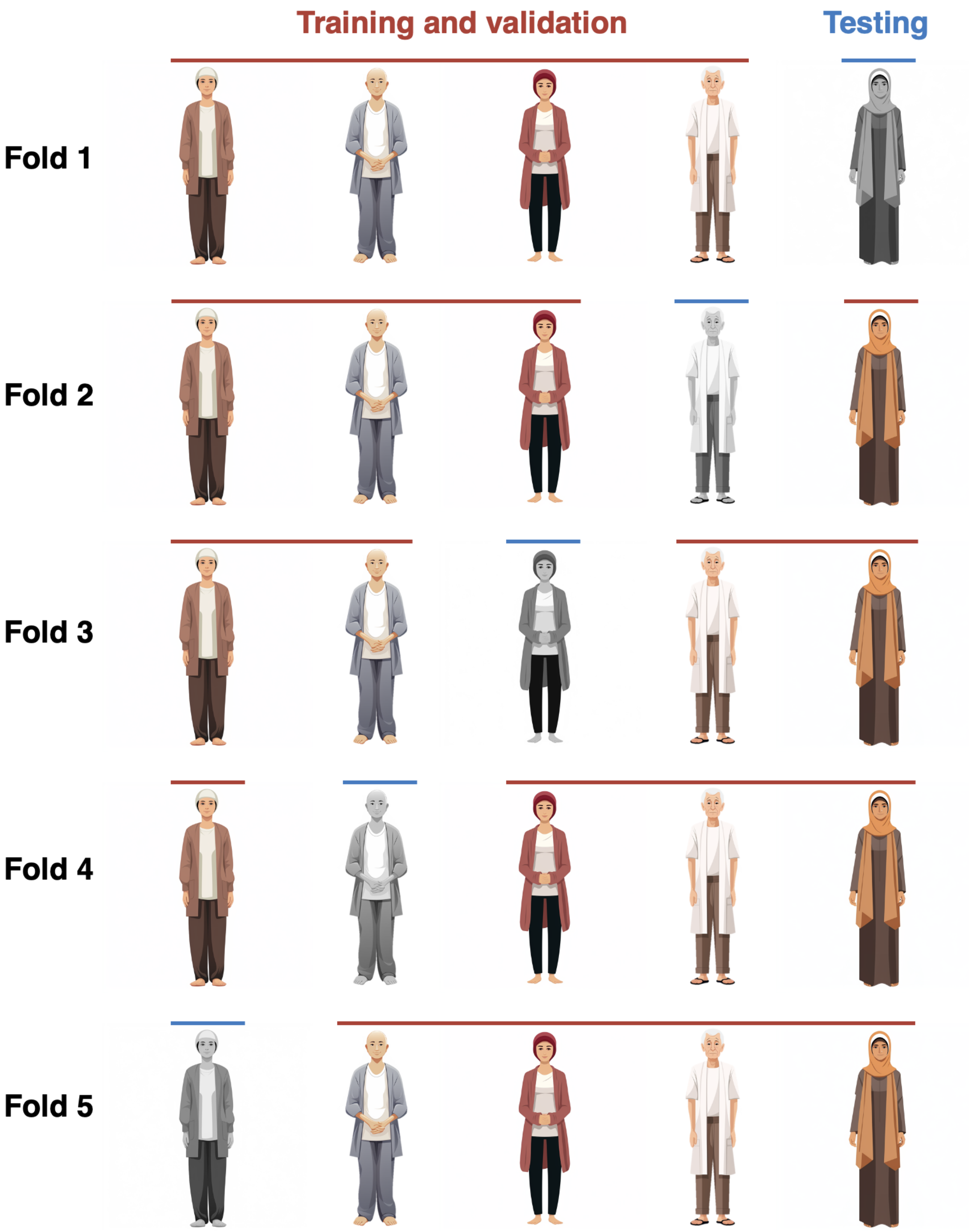

**Suppl. Figure 1: Visualization of 5-fold data splits.** A 5-fold data split leads to 5 permutations of a 80-20 split of the data. In other words, if you have 5 patients, one of the patients is left out for testing, and the remainder of patients are used to train the model. This is repeated 5 times, so that every patient has been in the training set 4 times, and in the testing set once. Consequently, training with a *k*-fold data split will result in *k* models trained with the same architecture, which have analyzed a slightly different subset of patients during training.